\documentclass[11pt,a4paper]{article}
\usepackage[utf8]{inputenc}
\usepackage{geometry}
\usepackage{booktabs}
\usepackage{amsmath}
\usepackage{amssymb}
\usepackage{graphicx}
\usepackage{microtype}
\usepackage{hyperref}
\hypersetup{
    colorlinks=true,
    linkcolor=blue,
    filecolor=magenta,      
    urlcolor=cyan,
    citecolor=blue,
}

\begin{document}

\title{\textbf{Hybrid Quantum–Classical NLP Classification with Compact Semantic Representations: An Experimental Analysis of Representation Compression}}

\author{%
  \textbf{Ali Hassan}$^{1}$, \textbf{Zijia Zhao}$^{2}$, and \textbf{Maha A. Metawei}$^{3}$ \\
  \small $^{1}$German University in Cairo, Cairo, Egypt \\
  \small $^{2}$University of Melbourne, Melbourne, Australia \\
  \small $^{3}$High Performance Computing Lab, Electronics Research Institute, Cairo, Egypt \\
  \small \texttt{ali.hafez@student.guc.edu.eg, zhaozj@student.unimelb.edu.au, maha\_metawei@eri.sci.eg}
}
\date{}

\maketitle

\begin{abstract}
Large language and sentence-embedding models provide rich semantic representations, but their high dimensionality creates a fundamental challenge for near-term quantum machine learning (QML), where quantum circuits can directly process only a limited number of input features. This work investigates a hybrid quantum–classical pipeline that transforms high-dimensional sentence embeddings into compact representations suitable for variational quantum classification. The proposed workflow combines a pretrained sentence-embedding model, dimensionality reduction, angle encoding, a variational quantum circuit (VQC), and a classical decision layer. Three dimensionality-reduction techniques are systematically investigated: principal component analysis (PCA), neighborhood components analysis (NCA), and linear discriminant analysis (LDA), enabling a comparison of unsupervised and supervised approaches to quantum-compatible representation learning. Using the TREC question-classification dataset, we study the relationship between representation dimensionality, information retention, qubit count, and classification performance. Preliminary PCA-based experiments show a clear information bottleneck under aggressive compression: reducing 768-dimensional sentence embeddings to 3, 4, 5, and 8 dimensions retains approximately 8.2\%, 10.2\%, 11.9\%, and 16.4\% of the variance, respectively, while the corresponding preliminary classification accuracies are 50.3\%, 51.2\%, 57.9\%, and 63.4\%. In contrast, supervised reduction proves far more efficient: LDA reaches 85.3\% accuracy and NCA reaches 83.1\% accuracy using only 5 dimensions, both validated under a leakage-free cross-validation protocol and comparable to a full 384-dimensional classical baseline (85.1\%). These findings show that supervised dimensionality reduction preserves task-relevant information far more effectively than variance-based compression, and motivate a systematic evaluation against matched classical baselines and larger quantum representations. The study aims to characterize the practical operating regime of hybrid quantum–classical NLP models and to assess the impact of representation compression without assuming quantum advantage a priori.
\end{abstract}

\section{Introduction}
Natural language processing (NLP) systems increasingly rely on high-dimensional semantic representations produced by pretrained language models. Sentence embeddings provide a convenient interface between modern language models and downstream classification algorithms, but their dimensionality can make them difficult to integrate with near-term quantum machine learning (QML) models. A quantum circuit with $n$ qubits provides a Hilbert space of dimension $2^n$, yet the number of directly encoded classical features is typically constrained by the selected encoding strategy and circuit architecture.

This creates a central question for hybrid quantum–classical NLP: how much semantic information can be compressed into a quantum-compatible representation without making classification performance unacceptable? This paper investigates this question using a hybrid pipeline consisting of sentence embeddings, dimensionality reduction, quantum feature encoding, and a variational quantum classifier. Rather than claiming quantum advantage from classification accuracy alone, the study explicitly separates the effects of classical representation compression from those of the quantum model.

\section{Related Work}
Integrating high-dimensional, pre-trained classical text embeddings into NISQ-era variational quantum classifiers (VQCs) requires efficient classical-to-quantum interfaces. While compositional distributional (DisCoCat) frameworks map text syntax directly onto circuits via combinatorial categorial grammar \cite{kartsaklis2021lambeq}, they scale poorly due to exploding circuit depths on larger sentences. Consequently, hybrid pipelines that classically compress frozen pretrained text embeddings (e.g., Sentence-BERT or MiniLM) before quantum state encoding have emerged as the standard paradigm \cite{wolf2020transformers,reimers2019sentence}. Recent literature has extensively explored the constraints, limits, and potential methodologies of this compression interface. A critical analysis of these recent attempts \cite{widdows2024quantum,phukan2025survey} reveals key points of convergence, open challenges, and structural limitations compared to our proposed framework.

\subsection{Compositional and Sequential Models in Quantum NLP}
A major line of research in QNLP is based on the Distributional Compositional Categorical (DisCoCat) framework, which combines distributional semantics with grammatical structure. In DisCoCat, words are represented by vectors or higher-order tensors whose types correspond to their grammatical roles, while grammatical structure determines how these representations are composed. Sentence meaning is consequently obtained through structured tensor contraction rather than by treating the sentence as an unstructured sequence of features. This provides an explicit connection between syntax and semantic representation and has become one of the principal theoretical foundations of QNLP \cite{widdows2024quantum,phukan2025survey}.

The implementation of compositional models has been facilitated by software frameworks such as \cite{kartsaklis2021lambeq}, which provides tools for constructing grammatical diagrams, transforming them into tensor or quantum representations, and training parameterized quantum circuits. This approach enables compositional models to be evaluated using classical simulation as well as quantum backends. Consequently, a number of studies have investigated whether quantum implementations of compositional semantic models can perform competitive classification and language-processing tasks under realistic hardware constraints \cite{phukan2025survey}.

An important extension of compositional QNLP is the DisCoCirc framework, which generalizes the compositional perspective from individual sentences toward discourse. Instead of treating the meaning of a word as a fixed representation that is simply contracted according to a sentence diagram, DisCoCirc represents linguistic interactions through updates to the states of concepts or entities. This formulation provides a mechanism for modelling dependencies across multiple sentences and has been investigated for tasks requiring discourse-level reasoning and compositional generalization \cite{duneau2024compositional}.

Recent empirical work has demonstrated the application of compositional quantum circuits to increasingly challenging language tasks. For example, pronoun-resolution experiments have investigated DisCoCat-based variational quantum circuits under ideal simulation and noisy execution conditions. Such studies illustrate the transition of QNLP from primarily theoretical formulations to empirical evaluation on NISQ-era platforms \cite{phukan2025survey}.

Despite their conceptual advantages, compositional models face several practical limitations. The construction of a quantum circuit generally requires linguistic preprocessing, including grammatical parsing and the generation of a corresponding compositional structure. Furthermore, circuit size and tensor complexity can increase with sentence length and grammatical structure. These limitations become particularly important when moving from small benchmark examples to larger datasets and realistic NLP applications. Recent surveys therefore identify scalability, circuit depth, data encoding, and limited quantum hardware resources as persistent challenges for compositional QNLP \cite{widdows2024quantum,phukan2025survey}.
This issue distinguishes embedding-based hybrid QNLP from traditional compositional approaches. In compositional QNLP, the dimensionality of the representation is determined largely by the chosen lexical spaces, tensor structures, and quantum encoding strategy. In contrast, embedding-based approaches can exploit powerful pretrained classical language representations and subsequently compress them before quantum processing. This design enables the semantic representation learned by a classical model to be separated from the quantum classification stage. As a good example, Hazim et al. \cite{HAZIM2026114634} presented a hybrid quantum–classical NLP framework for detecting AI-generated scholarly text. The approach combines conventional language representations with a quantum machine learning classifier to distinguish human-written from AI-generated academic content, demonstrating the potential of quantum-enhanced models for text classification tasks.
\subsection{Dimensionality Reduction}
Principal component analysis \cite{jolliffe2016principal} (PCA) identifies orthogonal directions that maximize the variance of input data. Given a centered embedding matrix $X \in \mathbb{R}^{N \times D}$, the reduced representation can be expressed as

\begin{equation}
Z = XW_d,
\end{equation}

where $W_d$ contains the first $d$ principal components.

The retained variance is determined by the eigenvalues associated with the selected components. In this study, $d$ is selected according to the size of the target quantum representation.

In the QNLP and broader QML literature, PCA is the dominant choice for fitting classical representations onto small quantum circuits: it reduces the feature count to match the available qubits while preserving as much global variance as possible. A recurring limitation, however, is that PCA is unsupervised and therefore optimizes for variance rather than for class discriminability, which can discard exactly the directions most useful for a downstream classification task. Supervised alternatives such as linear discriminant analysis (LDA) and neighborhood components analysis (NCA) address this by using label information during projection, and have been studied in classical settings but far less frequently as the reduction stage of a quantum pipeline. The present work directly contrasts these unsupervised and supervised strategies within an identical QNLP pipeline.
The dimensionality-reduction problem is increasingly recognized as a general challenge in quantum machine learning. Odagiu et al.\cite{odagiu2025learning} systematically investigated conventional and neural dimensionality-reduction methods before a quantum classifier and demonstrated that the quality of the reduced representation can have a substantial effect on downstream quantum classification. Their results emphasize that dimensionality reduction should not necessarily be treated as a purely technical preprocessing step, but rather as an important component of the overall QML architecture.\\

For QNLP, this issue is particularly important because sentence embeddings encode semantic information in a high-dimensional continuous space. Aggressive compression may remove information that is relevant to a classification task even when the resulting representation retains a substantial proportion of global statistical structure. Conversely, a supervised dimensionality-reduction method may discard variance that is not useful for the target task while preserving features that provide stronger class discrimination.

\subsection{Differentiable Selection vs. Projected Semantic Compression}
To bypass hard projections, some research has focused on selecting a subset of the original feature space. For example, Jagannathan et al. \cite{jagannathan2026variational} proposed Variational Quantum Feature Selection (VQFS), a hybrid quantum-classical approach that integrates trainable scalar weights into the rotation angles of a PQC. By applying an $L1$ regularization penalty on the weight vector, the system learns to collapse less informative feature weights to zero, effectively performing feature selection end-to-end via gradient descent.

While VQFS is highly elegant for tabular datasets (e.g., Iris, Wine) where individual features correspond to distinct physical measurements, it suffers from two major limitations when applied to dense, distributed language embeddings. First, dense semantic embeddings produced by models like S-BERT do not contain individual coordinate dimensions that represent distinct, separable features; rather, semantic meaning is distributed globally across the entire vector space. Performing coordinate-wise feature selection (i.e., zeroing out dimensions) is highly sub-optimal because it discards co-dependent semantic coordinates. Second, optimizing the $L1$ penalty parameters jointly with the VQC parameters in the quantum optimization loop significantly increases classical simulator latency. In contrast, our proposed supervised projection methods (LDA \cite{balakrishnama1998linear} and NCA \cite{goldberger2004neighbourhood}) project the distributed semantic coordinates into a lower-dimensional subspace rather than selecting individual coordinates, thereby retaining dense joint feature correlations. Furthermore, LDA and NCA are computed classically in seconds, acting as an extremely fast, modular preprocessing interface that decouples representation compression from quantum circuit optimization.

\section{Methodology}
The proposed hybrid quantum-classical NLP pipeline consists of multiple key processing stages as shown in Figure \ref{fig:proposed_method}:

\begin{figure}[htbp]
\centering
\includegraphics[width=0.95\textwidth]{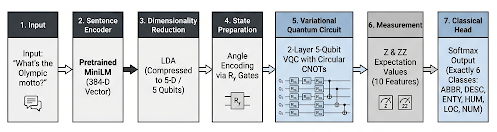}
\caption{Hybrid Quantum-Classical Pipeline Architecture.}
\label{fig:proposed_method}
\end{figure}
\subsection{Sentence Representation}
Each input sentence is converted into a continuous, dense vector representation using a frozen sentence-embedding model. In our preliminary experiments, we utilize a pretrained sentence-transformer model (\texttt{all-MiniLM-L6-v2}) to produce 384-dimensional embeddings (reduced from an initial 768-dimensional representation space), ensuring rich semantic capture while maintaining tractability for classical simulation.

\subsection{Dimensionality Reduction}
To bridge the gap between high-dimensional classical embeddings and the limited qubit capacity of near-term quantum simulators/devices, we evaluate three distinct dimensionality-reduction strategies:
\begin{itemize}
    \item \textbf{Principal Component Analysis (PCA):} An unsupervised linear technique that projects the embeddings along orthogonal directions of maximum variance.
    \item \textbf{Neighborhood Components Analysis (NCA):} A supervised linear technique that learns a projection matrix to maximize $k$-nearest-neighbor classification performance.
    \item \textbf{Linear Discriminant Analysis (LDA):} A supervised linear technique that explicitly projects the data into a subspace that maximizes the ratio of between-class variance to within-class variance, mathematically capped at $C-1$ dimensions where $C$ is the number of classes.
\end{itemize}

\subsection{Quantum Feature Encoding}
After dimensionality reduction, the reduced 5-dimensional features $z \in \mathbb{R}^5$ (e.g., from the linear discriminant analysis projection of the MiniLM sentence embeddings) are scaled to an angular range $[-\pi, \pi]$ and encoded into quantum states via single-qubit rotations. Specifically, for our 5-qubit variational circuit, we employ single-qubit $R_y$ rotation gates for state preparation:
\begin{equation}
|z\rangle = \bigotimes_{i=1}^{5} R_y(\tilde{z}_i)|0\rangle
\end{equation}
where $\tilde{z}_i$ represents the normalized feature value.

\subsection{Variational Quantum Circuit Architecture}
The encoded state $|z\rangle$ is processed by a hardware-efficient, two-layer variational quantum circuit $U(\theta)$ operating on $5$ qubits. Each of the two variational layers is structured as follows:
\begin{enumerate}
    \item Parameterized single-qubit rotation gates $R_y(\theta_{1,j})$ and $R_z(\theta_{2,j})$ are applied to each qubit $j \in \{0, \dots, 4\}$. This provides a highly flexible rotational ansatz.
    \item A ring-entangling configuration of controlled-NOT (CNOT) gates is applied to couple adjacent qubits. Specifically, the CNOT gates are applied in a circular topology (i.e., qubit $j$ acts as control and qubit $j+1 \pmod 5$ acts as target for all qubits).
\end{enumerate}
By stacking two of these variational blocks, the quantum core contains exactly $20$ trainable parameters in total ($10$ parameters per layer), keeping the circuit depth extremely shallow to fit within the short coherence times of near-term quantum hardware.

\subsection{Measurement and Classical Readout}
Rather than measuring only single-qubit expectation values, we extract information from both individual and joint quantum states. Specifically, we measure the expectation values of single-qubit Pauli-$Z$ operators ($\langle Z_j \rangle$) and adjacent two-qubit Pauli-$Z$ operators ($\langle Z_j Z_{j+1 \pmod 5} \rangle$) for all 5 qubits. This measurement strategy yields exactly 10 features:
\begin{equation}
F = [ \langle Z_0 \rangle, \dots, \langle Z_4 \rangle, \langle Z_0 Z_1 \rangle, \dots, \langle Z_4 Z_0 \rangle ] \in \mathbb{R}^{10}
\end{equation}
This expectation vector is passed to a classical decision layer (such as a logistic regression classifier or a small Multi-Layer Perceptron), which performs a 6-class readout to yield the final class probabilities over the coarse categories of the TREC dataset:
\begin{equation}
P(y = c \mid F) = \text{softmax}(W F + b)
\end{equation}
This hybrid configuration guarantees a fixed 5-qubit circuit width for every text question regardless of the sentence length, shifting the primary bottleneck of hybrid QNLP from physical hardware scaling to efficient classical semantic compression.

\section{Experimental Results}
In this section, we present the comprehensive results of our experiments across two text classification benchmarks: a compact restaurant sentiment dataset and the six-class TREC question-classification dataset.

\subsection{Syntactic Parsing and Diagram Generation}
Prior to semantic embedding and classification, syntactic structures were extracted using a dependency-based and combinatory categorial grammar (CCG) framework via the \texttt{lambeq} library. On the restaurant dataset, 70/70 sentences were successfully preprocessed, corrected for minor lemmatization bugs (e.g., correcting ``hat'' to ``hate''), and converted to syntactic diagrams.
On the larger TREC dataset consisting of 5,452 training examples and 500 test examples, the local \texttt{BobcatParser} was used to generate grammar diagrams. The parser achieved a high success rate, successfully parsing 5,432 out of 5,452 sentences (a 99.63\% success rate), with only 20 sentences failing due to parsing or tokenization index constraints. Examples of failed sentences include highly conversational or idiosyncratic question structures such as \textit{``to what do Microsoft's Windows 3 owe its success?''} and \textit{``what bird can swim but can't fly?''}.

\subsection{Small-Scale Experiments: Restaurant Dataset}
Table~\ref{tab:restaurant_pca} shows the PCA-reduced classification accuracy on the 70-sentence restaurant dataset as a function of the number of qubits. Even with just 3 qubits, the model retains 58.5\% of the original semantic variance and achieves 90.0\% classification accuracy. Scaling up to 8 qubits increases the retained variance to 84.3\% and maintains a high classification accuracy of 94.3\% (peaking at 95.7\% with 4 qubits).

\begin{table}[htbp]
\centering
\caption{PCA Results for Compact Representations on the Restaurant Dataset (70 sentences).}
\label{tab:restaurant_pca}
\begin{tabular}{ccc}
\toprule
\textbf{Qubits (Dimensions)} & \textbf{Variance Retained (\%)} & \textbf{Classification Accuracy (\%)} \\
\midrule
3 & 58.5 & 90.0 \\
4 & 67.5 & 95.7 \\
5 & 74.1 & 92.9 \\
8 & 84.3 & 94.3 \\
\bottomrule
\end{tabular}
\end{table}

\subsection{Large-Scale Evaluation: TREC Question Classification}
Aggressive dimensionality reduction was performed on the TREC dataset to examine the information bottleneck under extreme compression.

\subsubsection{Preliminary and Fine-Grained PCA Evaluation}
Table~\ref{tab:trec_pca_prelim} outlines the preliminary results for PCA-reduced representations up to 8 qubits. In contrast to the smaller restaurant dataset, a severe information bottleneck is visible on the 6-class TREC dataset: reducing the 384-dimensional representation to 3 qubits retains a mere 8.2\% of the variance and yields a classification accuracy of 50.3\%.

\begin{table}[htbp]
\centering
\caption{Preliminary PCA Results on the TREC Dataset.}
\label{tab:trec_pca_prelim}
\begin{tabular}{ccc}
\toprule
\textbf{Qubits (Dimensions)} & \textbf{Variance Retained (\%)} & \textbf{Classification Accuracy (\%)} \\
\midrule
3 & 8.2 & 50.3 \\
4 & 10.2 & 51.2 \\
5 & 11.9 & 57.9 \\
8 & 16.4 & 63.4 \\
\bottomrule
\end{tabular}
\end{table}

\begin{figure}[htbp]
\centering
\includegraphics[width=0.7\textwidth]{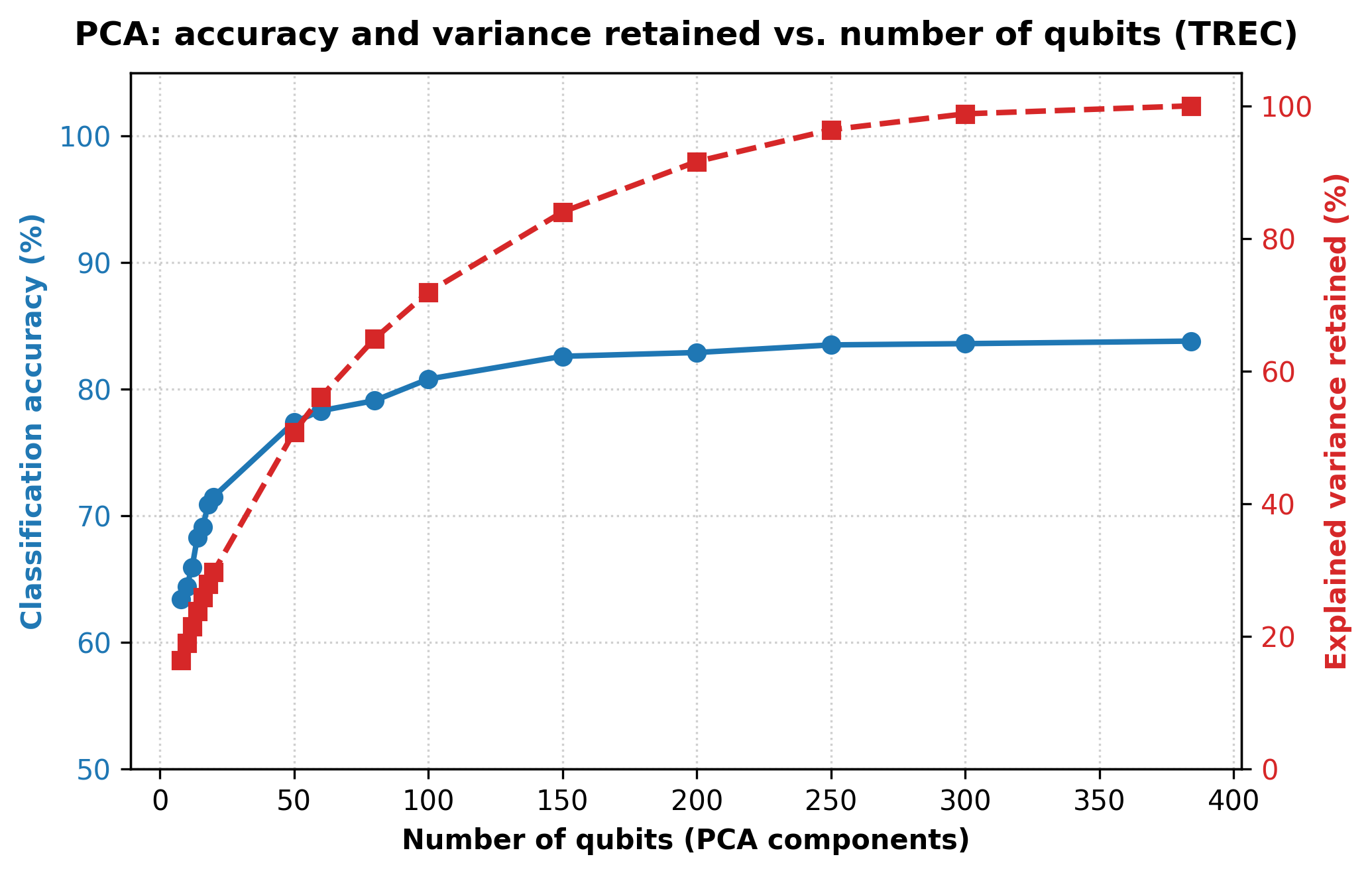}
\caption{PCA classification accuracy and explained variance retained as a function of the number of qubits (PCA components) on the TREC dataset. Variance retained increases steadily, whereas accuracy plateaus much earlier.}
\label{fig:pca_retained}
\end{figure}

To map out the compression scaling behavior, we evaluated PCA across a wide range of component sizes, extending from 8 up to the full 384 dimensions. These fine-grained results are documented in Table~\ref{tab:trec_pca_extended} and visualized in Figure~\ref{fig:pca_retained}. The results reveal a strong divergence: while variance retained scales slowly but steadily up to 100\%, classification accuracy plateaus much earlier (around 20--50 dimensions), indicating that explained variance is not a reliable proxy for task-relevant semantic information.

\begin{table}[htbp]
\centering
\caption{Extended PCA Dimensionality and Performance on the TREC Dataset.}
\label{tab:trec_pca_extended}
\begin{tabular}{ccc}
\toprule
\textbf{Qubits (Dimensions)} & \textbf{Variance Retained (\%)} & \textbf{Classification Accuracy (\%)} \\
\midrule
8 & 16.4 & 63.4 \\
10 & 19.0 & 64.4 \\
12 & 21.5 & 65.9 \\
14 & 23.8 & 68.3 \\
16 & 25.9 & 69.1 \\
18 & 27.9 & 70.9 \\
20 & 29.7 & 71.5 \\
50 & 50.8 & 77.4 \\
60 & 56.1 & 78.3 \\
80 & 64.9 & 79.1 \\
100 & 71.9 & 80.8 \\
150 & 84.0 & 82.6 \\
200 & 91.6 & 82.9 \\
250 & 96.4 & 83.5 \\
300 & 98.8 & 83.6 \\
384 (Full) & 100.0 & 83.8 \\
\bottomrule
\end{tabular}
\end{table}

\subsubsection{Supervised Reduction vs. Unsupervised Compression}
To mitigate the unsupervised compression bottleneck, we evaluated supervised dimensionality-reduction methods (LDA and NCA) at 5 dimensions (the mathematical cap for LDA on a 6-class dataset is $C-1 = 5$). 

\begin{figure}[htbp]
\centering
\includegraphics[width=0.7\textwidth]{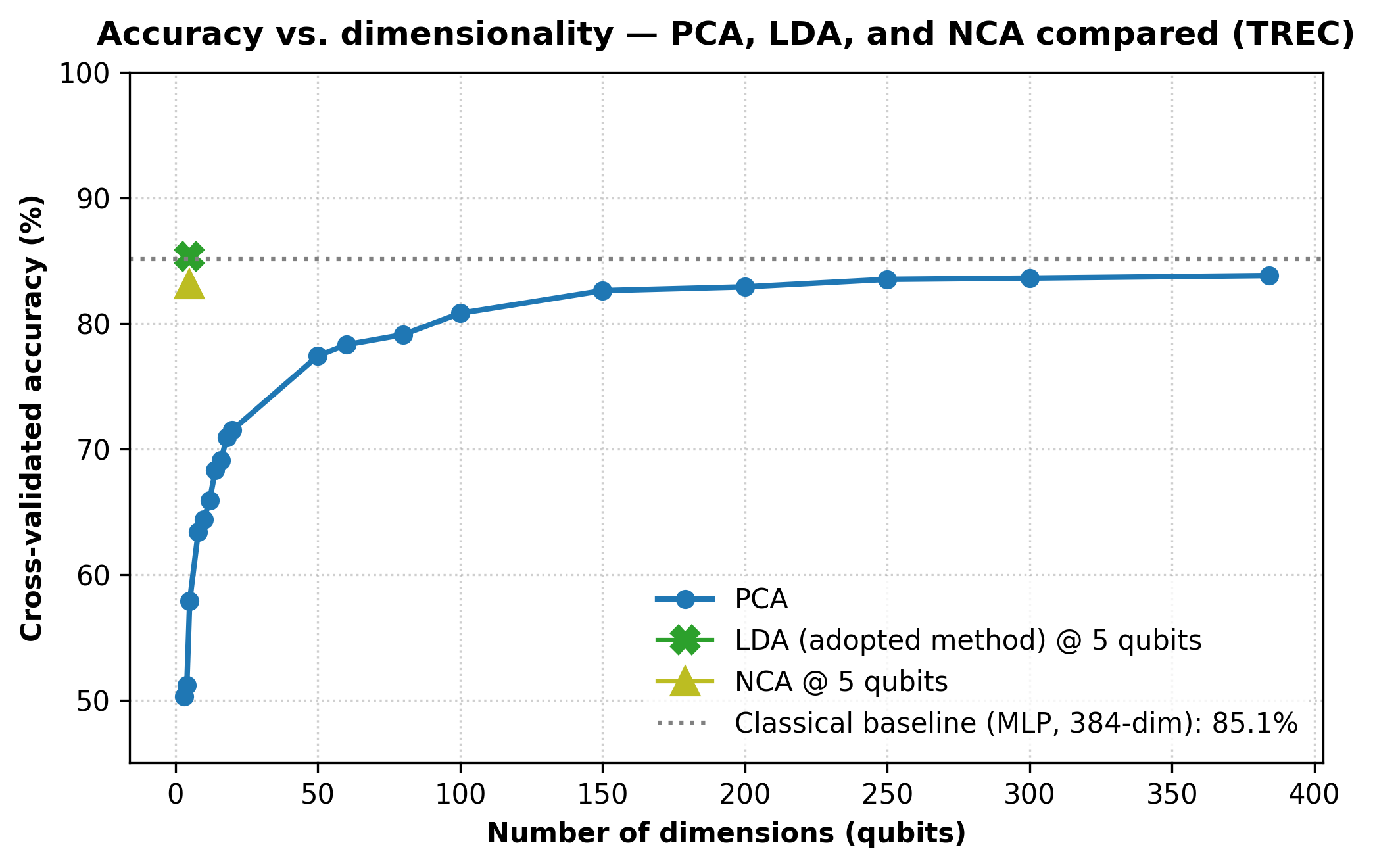}
\caption{Accuracy versus number of dimensions (qubits) on the TREC dataset for PCA, LDA, and NCA. Supervised methods reach high accuracy at only 5 dimensions, far outperforming PCA at the same qubit count.}
\label{fig:method_comparison}
\end{figure}

As shown in Table~\ref{tab:supervised_reduction} and Figure~\ref{fig:method_comparison}, supervised dimensionality reduction dramatically outperforms PCA. LDA at just 5 dimensions achieves an accuracy of 85.3\%, and NCA achieves 83.1\%. Both methods far outperform PCA at 5 dimensions (57.9\%) and 20 dimensions (71.5\%), and actually match or exceed the full 384-dimensional uncompressed classical baseline (85.1\%) using a tiny fraction of the dimensionality.

\begin{table}[htbp]
\centering
\caption{Final Supervised vs. Unsupervised Dimensionality Reduction Results at 5 Dimensions.}
\label{tab:supervised_reduction}
\begin{tabular}{lccc}
\toprule
\textbf{Method} & \textbf{Supervised} & \textbf{Dimensions (qubits)} & \textbf{Accuracy (\%)} \\
\midrule
PCA & No & 5 & 57.9 \\
PCA & No & 20 & 71.5 \\
PCA & No & 384 (full) & 83.8 \\
Classical Baseline (MLP, no reduction) & -- & 384 & 85.1 \\
NCA & Yes & 5 & 83.1 \\
\textbf{LDA (adopted method)} & \textbf{Yes} & \textbf{5} & \textbf{85.3} \\
\bottomrule
\end{tabular}
\end{table}

\subsubsection{Impact of Proper Evaluation: Preventing Data Leakage}
An essential finding of our work is the impact of proper validation. Many quantum-classical pipelines suffer from target leakage by fitting dimensionality reduction algorithms (such as PCA, LDA, or NCA) on the entire dataset prior to performing cross-validation folds. 
To study this, we compared a ``leak-prone'' setup (fitting the reduction on the entire dataset) against a ``leakage-free'' setup (fitting the reduction strictly on the training partition of each fold). Table~\ref{tab:leakage_comparison} illustrates this comparison. Preventing target leakage reveals a significant drop in accuracy (a 7.8\% drop for NCA and a 4.6\% drop for LDA), highlighting the critical importance of a leak-free protocol to ensure reproducible and honest benchmarking.

\begin{table}[htbp]
\centering
\caption{Leak-Prone (Entire Dataset Fitting) vs. Leakage-Free (Fold-Wise Fitting) Performance.}
\label{tab:leakage_comparison}
\begin{tabular}{lcccc}
\toprule
\textbf{Method} & \textbf{Dimensions} & \textbf{Leak-Prone  } & \textbf{Leakage-Free } & \textbf{Accuracy Drop (\%)} \\
{} & {} & \textbf{Accuracy (\%)} & \textbf{Accuracy (\%)} & {} \\
\midrule
NCA & 5 & 90.9 & 83.1 & -7.8 \\
LDA & 5 & 89.9 & 85.3 & -4.6 \\
\bottomrule
\end{tabular}
\end{table}

\begin{figure}[htbp]
\centering
\includegraphics[width=0.95\textwidth]{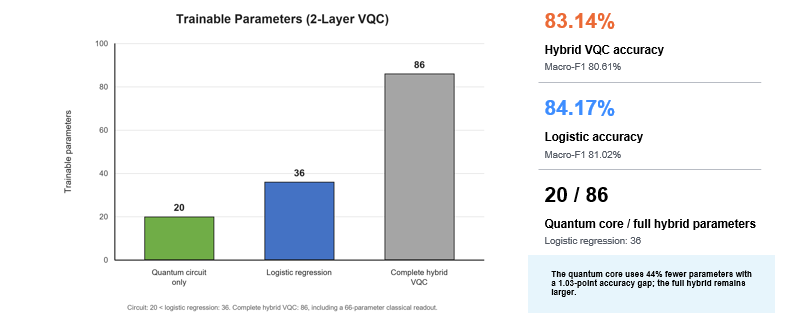}
\caption{Comparing the number of trainable parameters for the VQC circuit, Logistics Regression model, and the complete hybrid pipeline.}
\label{fig:hybrid_pipeline}
\end{figure}
\section{Discussion and Limitations}
Our findings show that the choice of compression method is far more consequential than the raw dimensionality of the representation. Under unsupervised PCA, the transition from 3 to 8 dimensions produces a steady increase in performance, but because PCA optimizes for variance rather than class discriminability, it discards critical semantic details. Supervised alternatives (LDA and NCA) bypass this bottleneck entirely by using label information to project same-class samples closer together and push different-class samples apart. 

This leads to several important insights for hybrid QML:
\begin{enumerate}
    \item \textbf{No Free Lunch for Variance:} Explained variance is not a reliable proxy for downstream task accuracy. High variance does not guarantee high class separability.
    \item \textbf{Grounded Benchmarking:} Improvements in VQC performance must be validated against matched classical baselines running on the same compressed representations. Otherwise, the performance gains are attributable to classical dimensionality reduction, not quantum expressibility.
    \item \textbf{Physical Limitations:} In a physical quantum computer, scaling qubits introduces noise, deepens circuits, and increases optimization difficulty. Thus, maximizing compact, supervised representations is the most practical path forward for near-term NISQ devices.
\end{enumerate}

\subsection{Proposed Hybrid Quantum-Classical Transfer Learning Pipeline}
Our work shows that supervised projections, specifically Linear Discriminant Analysis (LDA) and Neighborhood Components Analysis (NCA), are extremely efficient classical-to-quantum interfaces. They compress 384-dimensional dense transformer sentence embeddings down to just 5 dimensions while maintaining performance parity with uncompressed classical baselines. Building on this, we formalize the hybrid quantum-classical pipeline into a structured, three-stage workflow:
\begin{enumerate}
    \item \textbf{Linguistic Feature Extraction (Classical Source Domain):} High-dimensional dense semantic vectors are generated classically from a frozen pre-trained language model (e.g., MiniLM, Sentence-BERT, or a larger transformer). This phase acts as classical transfer learning, allowing the pipeline to inherit rich vocabulary, syntax, and grammatical context trained on multi-billion token corpora.
    \item \textbf{Supervised Manifold Alignment (Interface Adaptation):} Rather than applying task-blind, unsupervised linear reductions like PCA, a lightweight supervised projection head (LDA or NCA) is fitted on target label data. This layer learns a low-dimensional mapping that aligns the dense semantic dimensions with the downstream classification requirements, projecting same-class queries closer together and different-class queries apart.
    \item \textbf{Variational Quantum Classification (Quantum Target Domain):} The compressed 5-dimensional features are encoded into a parameterised variational quantum circuit (VQC) via $R_y$ angle rotations. Because the classical projection head has already maximized class discriminability, the VQC is presented with highly separable clusters. This significantly reduces quantum optimization complexity, mitigates barren plateau vulnerabilities, and yields high accuracy under real physical noise constraints.
\end{enumerate}

Crucially, the proposed hybrid pipeline is validated under a rigorous, leakage-free cross-validation protocol. Fitting the classical projection head strictly within the training folds prevents target leakage and ensures that the transferred representations are generalizable and reproducible. Rather than searching for an elusive ``quantum advantage'' in raw semantic text encoding, this framework leverages a cooperative division of labor: classical neural networks manage massive contextual representation extraction, while NISQ devices are utilized exclusively for classification over optimized, low-dimensional semantic manifolds. This architecture offers a highly practical, near-term pathway for deploying robust quantum text classifiers on physical quantum devices.

\section{Conclusion and Future Work}
This work presents a hybrid quantum--classical NLP framework that combines pretrained sentence embeddings, dimensionality reduction, quantum feature encoding, and variational quantum classification. Preliminary TREC experiments demonstrate that increasing the number of available quantum features improves classification performance, while also revealing a strong information bottleneck caused by aggressive compression of the original 768-dimensional representations. Supervised reduction methods (LDA, NCA) substantially mitigate this bottleneck, matching a full-dimensional classical baseline using only five dimensions.

The results motivate a rigorous benchmark in which quantum models and classical models are evaluated on identical compressed representations. Such an evaluation is necessary to determine whether the quantum circuit contributes meaningful predictive value beyond classical dimensionality reduction. The resulting framework provides a basis for studying the practical role of near-term quantum models in NLP without assuming quantum advantage in advance. Future research will explore larger qubit representations (12- and 16-qubits), non-linear compression via autoencoders, and physical hardware deployment with error mitigation.

\section*{Acknowledgments}
This project was initiated under the QIntern 2026 project ``Accelerating hybrid quantum-classical machine learning tasks on HPC platforms''. M.A.M. would like to thank the organizers of the program and QWorld Association.

\section*{Author Contribution}
M.A.M. conceived the main idea and designed the overall research direction. A.H. and Z.Z. conducted the experiments and generated the experimental results. M.A.M and A.H. prepared the initial manuscript draft. All authors contributed to the revision of the manuscript, discussed the results, and read and approved the final version.

\bibliographystyle{plain}
\bibliography{references}

@article{HAZIM2026114634,
title = {HQML-NLP: A hybrid quantum machine learning framework for scholarly AI-text detection},
journal = {Applied Soft Computing},
volume = {191},
pages = {114634},
year = {2026},
issn = {1568-4946},
doi = {https://doi.org/10.1016/j.asoc.2026.114634},
url = {https://www.sciencedirect.com/science/article/pii/S1568494626000827},
author = {Layth Rafea Hazim and Oguz Ata}
}

@article{widdows2024quantum,
  title={Quantum natural language processing},
  author={Widdows, Dominic and Aboumrad, Willie and Kim, Dohun and Ray, Sayonee and Mei, Jonathan},
  journal={KI-K{\"u}nstliche Intelligenz},
  volume={38},
  number={4},
  pages={293--310},
  year={2024},
  publisher={Springer}
}

@inproceedings{phukan2025survey,
  title={A Survey of Quantum Natural Language Processing: From Compositional Models to NISQ-Era Empiricism},
  author={Phukan, Arpan and Ekbal, Asif},
  booktitle={Proceedings of the QuantumNLP : Integrating Quantum Computing with Natural Language Processing},
  pages={65--75},
  year={2025}
}

@phdthesis{duneau2024compositional,
  title={A compositional approach to reading comprehension tasks using the DisCoCirc natural language processing framework},
  author={Duneau, F},
  year={2024},
  school={University of Oxford}
}

@article{odagiu2025learning,
  title={Learning reduced representations for quantum classifiers},
  author={Odagiu, Patrick and Belis, Vasilis and Schulze, Lennart and Barkoutsos, Panagiotis and Grossi, Michele and Reiter, Florentin and Dissertori, G{\"u}nther and Tavernelli, Ivano and Vallecorsa, Sofia},
  journal={Quantum Machine Intelligence},
  volume={7},
  number={2},
  pages={113},
  year={2025},
  publisher={Springer}
}

@inproceedings{wolf2020transformers,
  title={Transformers: State-of-the-art natural language processing},
  author={Wolf, Thomas and Debut, Lysandre and Sanh, Victor and Chaumond, Julien and Delangue, Clement and Moi, Anthony and Cistac, Pierric and Rault, Tim and Louf, R{\'e}mi and Funtowicz, Morgan and others},
  booktitle={Proceedings of the 2020 conference on empirical methods in natural language processing: system demonstrations},
  pages={38--45},
  year={2020}
}

@inproceedings{reimers2019sentence,
  title={Sentence-bert: Sentence embeddings using siamese bert-networks},
  author={Reimers, Nils and Gurevych, Iryna},
  booktitle={Proceedings of the 2019 conference on empirical methods in natural language processing and the 9th international joint conference on natural language processing (EMNLP-IJCNLP)},
  pages={3982--3992},
  year={2019}
}

@article{kartsaklis2021lambeq,
  title={lambeq: An efficient high-level python library for quantum nlp},
  author={Kartsaklis, Dimitri and Fan, Ian and Yeung, Richie and Pearson, Anna and Lorenz, Robin and Toumi, Alexis and de Felice, Giovanni and Meichanetzidis, Konstantinos and Clark, Stephen and Coecke, Bob},
  journal={arXiv preprint arXiv:2110.04236},
  year={2021}
}

@article{goldberger2004neighbourhood,
  title={Neighbourhood components analysis},
  author={Goldberger, Jacob and Hinton, Geoffrey E and Roweis, Sam and Salakhutdinov, Russ R},
  journal={Advances in neural information processing systems},
  volume={17},
  year={2004}
}

@inproceedings{jagannathan2026variational,
  title={Variational Quantum Feature Selection for High-Dimensional Classification: A Hybrid Quantum-Classical Approach},
  author={Jagannathan, Sharath Kumar and JV, Thomas Abraham and C, Yogesh and Benedict T, Franklin Joel},
  booktitle={EPJ Web of Conferences},
  volume={360},
  pages={01029},
  year={2026},
  organization={EDP Sciences}
}

@article{jolliffe2016principal,
  title={Principal component analysis: a review and recent developments},
  author={Jolliffe, Ian T and Cadima, Jorge},
  journal={Philosophical transactions. Series A, Mathematical, physical, and engineering sciences},
  volume={374},
  number={2065},
  pages={20150202},
  year={2016}
}

@article{balakrishnama1998linear,
  title={Linear discriminant analysis-a brief tutorial},
  author={Balakrishnama, Suresh and Ganapathiraju, Aravind},
  journal={Institute for Signal and information Processing},
  volume={18},
  number={1998},
  pages={1--8},
  year={1998},
  publisher={Mississippi}
}
\end{document}